# Distributed Fine-Grained Traffic Speed Prediction for Large-Scale Transportation Networks based on Automatic LSTM Customization and Sharing

Ming-Chang Lee[1], Jia-Chun Lin[2], and Ernst Gunnar Gran[3]

*[1,2,3]Department of Information Security and Communication Technology, Norwegian University of Science and Technology,*
*Ametyst-bygget, 2815 Gjøvik, Norway*
*[3]Simula Research Laboratory, 1364 Fornebu, Norway*

*[1] ming-chang.lee@ntnu.no*
*[2]jia-chun.lin@ntnu.no*
*[3]ernst.g.gran@ntnu.no*

3[rd] Jun 2020

# Distributed Fine-Grained Traffic Speed Prediction for Large-Scale Transportation Networks based on Automatic LSTM Customization and Sharing

Ming-Chang Lee[1], Jia-Chun Lin[2], and Ernst Gunnar Gran[3,4]

[1,2,3]Department of Information Security and Communication Technology, Norwegian University of Science and Technology, 2815 Gjøvik, Norway
[4]Simula Research Laboratory, 1364 Fornebu, Norway
[1]`ming-chang.lee@ntnu.no`
[2]`jia-chun.lin@ntnu.no`
[3]`ernst.g.gran@ntnu.no`

**Abstract.** Short-term traffic speed prediction has been an important research topic in the past decade, and many approaches have been introduced. However, providing fine-grained, accurate, and efficient traffic-speed prediction for large-scale transportation networks where numerous traffic detectors are deployed has not been well studied. In this paper, we propose DistPre, which is a distributed fine-grained traffic speed prediction scheme for large-scale transportation networks. To achieve fine-grained and accurate traffic-speed prediction, DistPre customizes a Long Short-Term Memory (LSTM) model with an appropriate hyperparameter configuration for a detector. To make such customization process efficient and applicable for large-scale transportation networks, DistPre conducts LSTM customization on a cluster of computation nodes and allows any trained LSTM model to be shared between different detectors. If a detector observes a similar traffic pattern to another one, DistPre directly shares the existing LSTM model between the two detectors rather than customizing an LSTM model per detector. Experiments based on traffic data collected from freeway I5-N in California are conducted to evaluate the performance of DistPre. The results demonstrate that DistPre provides time-efficient LSTM customization and accurate fine-grained traffic-speed prediction for large-scale transportation networks.

## 1    Introduction

Accurate traffic-speed prediction is crucial to achieve efficient proactive traffic management and control for large-scale transportation networks. During the past decade, many approaches and methods have been introduced for short-term traffic speed prediction. They can be classified into two categories: parametric approaches and non-parametric approaches. The former category of approaches simplifies the mapping function to a known form, i.e., these approaches require a pre-defined model. A typical example is the autoregressive integrated moving average approach (ARIMA) [1]. On the other hand, the nonparametric approaches make no assumptions about the form of the mapping function, i.e., they require no pre-defined model structure. The k-nearest neighbors (k-NN) method [2-3], artificial neural network (ANN) [4], recurrent neural network (RNN) [5], etc., all belong to this category. As a special type of RNN, long short-term memory [6], abbreviated as LSTM, is superior in time series prediction with long temporal dependencies. Prior studies such as [7-9] have proven that LSTM provides better prediction accuracy than many other approaches and neural networks. Therefore, LSTM is chosen as a building block for traffic speed prediction in this paper.

However, several issues still need to be addressed to achieve fine-grained, accurate, and efficient traffic speed prediction for large-scale transportation networks. For example, in large-scale transportation networks, numerous detectors, such as loop detectors or traffic cameras, are deployed in different places to collect traffic data. Depending on the density of nearby population and other factors, the traffic observed/collected by detectors at different locations may have diverse patterns. For instance, Fig. 1 shows that five detectors deployed on freeway I5-N in California [10] observe completely different traffic patterns between 6 a.m. and 10 a.m. in a typical weekday. In order to provide fine-grained traffic-speed prediction for achieving better transportation services and management, we suggest that each detector should have its own LSTM model to predict the traffic speed of its coverage. However, such an approach would be expensive, time consuming, and impractical because we might need to manually configure LSTM hyperparameters and train the corresponding LSTM model for each individual detector for several times until we find an LSTM model that is able to accurately make prediction. Note that LSTM hyperparameters are parameters whose values are set before the training process of an LSTM starts.

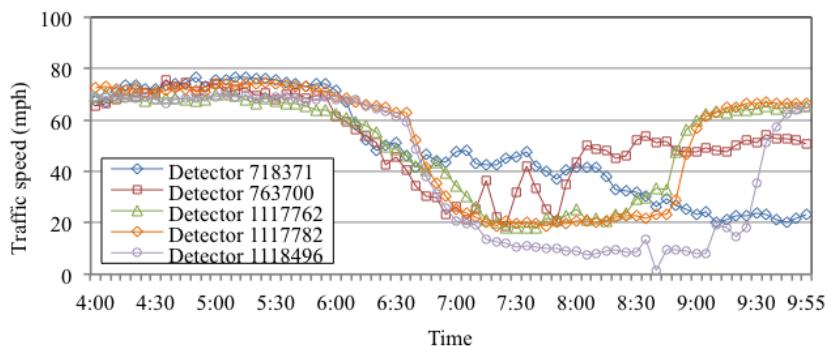

**Fig. 1.** The traffic speed collected by five randomly-chosen detectors on freeway I5-N in California between 4 a.m. and 10 a.m. in a typical weekday.

To address the above issue, we propose DistPre, which is a distributed fine-grained traffic speed prediction scheme for large-scale transportation networks. DistPre cus-tomizes an LSTM model for a detector by automatically determining LSTM hyperparameter values and training the corresponding LSTM model based on the Nelder-Mead method [11], which is a commonly applied method used to find the minimum or maximum of an objective function in a multidimensional space. To make the above customization process time-efficient for large-scale transportation networks, we propose that detectors should share the same LSTM model if they observe similar traffic patterns. More specifically, DistPre works in an incremental, distributed, and parallel manner. Whenever DistPre encounters an unprocessed detector $i$, it checks if the traffic-speed pattern observed by detector $i$ is similar to the one observed by any other detector that has been processed by DistPre. If the answer is negative, DistPre requests an available compute node from a computer cluster to customize an LSTM model for detector $i$. However, if the traffic-speed pattern observed by detector $i$ is similar to the one observed by detector $j$, DistPre directly shares the LSTM of detector $j$ with detector $i$ without requiring to customize a new LSTM model for detector $i$.

To demonstrate the performance of DistPre, we conducted experiments on an Apache Hadoop YARN cluster using real-world traffic data collected by detectors on freeway I5-N in California. The results confirm that DistPre is able to provide fine-grained and accurate traffic speed prediction for large-scale transportation networks due to the LSTM customization. In addition, DistPre is scalable, efficient, and cost-effective since the number of LSTM models does not proportionally increase with the number of detectors, due to the LSTM sharing feature of DistPre.

The rest of the paper is organized as follows: Section 2 briefly introduce LSTM, LSTM hyperparameters, and the Nelder-Mead method. Section 3 presents related work, while Section 4 introduces the details of DistPre. In Section 5, we evaluate the performance of DistPre. Section 6 concludes this paper and outlines future work.

## 2    LSTM, LSTM hyperparameters, and the Nelder-Mead Method

In this section, we introduce LSTM, LSTM hyperparameters, and the Nelder-Mead method.

### 2.1    LSTM and LSTM Hyperparameters

LSTM [6] is designed to learn long-term dependencies and model temporal sequences. The architecture of LSTM is similar to that of RNN except that the nonlinear units in the hidden layers are memory blocks. Each block contains memory cells, an input gate, an output gate, and a forget gate. The input gate decides whether the input should be stored in the memory cells or not. The output gate determines if current memory contents should be output. The forget gate decides if current memory contents should be erased. These features enable LSTM to preserve information over long time lags, thus addressing the vanishing gradient problem [12].

It is well-known that the prediction performance of LSTM highly depends on choosing appropriate values for the following hyperparameters:

- Learning rate (denoted by $R_{Learn}$)
- The number of hidden layers (denoted by $N_{Layer}$)
- The number of hidden units (denoted by $N_{Unit}$)
- Epochs (denoted by $ep$)

The **learning rate** controls how much the weights of LSTM are adjusted with respect to the loss gradient. The lower the value, the less is the chance to miss any local minima, but it prolongs the training process. A **hidden layer** is a layer between the input layer of LSTM and the output layer of LSTM. The more complex the training dataset is, the more hidden layers are required to learn the training dataset. A **hidden unit** is a neuron in a hidden layer. It is responsible for taking in a set of weighted inputs and produce an output through an activation function. Too many hidden units may result in overfitting, while too few hidden units might cause underfitting. An **epoch** is defined as one forward pass and one backward pass of all the training data. Too many epochs might overfit the training data, whereas too few epochs may underfit the training data.

Due to the importance of the above-mentioned hyperparameters to the learning performance and computational efficiency of LSTM, this paper takes all of them into consideration. One of this paper's goals is to automatically determine appropriate values for these hyperparameters such that the resulting LSTM model is able to achieve high prediction accuracy and that human effort can be greatly reduced.

## 2.2 The Nelder-Mead Method (NMM)

NMM [11] is a popular optimization method for non-linear functions. In this paper, we use it to automatically find appropriate values for the above-mentioned LSTM hyperparameters. NMM minimizes the target objective function by generating an initial simplex based on a predefined vertex and then performing a function evaluation at each vertex of the simplex. Note that a simplex has $n + 1$ vertices in $\mathbb{R}^n$ where $n$ is the number of dimensions of the parameter space. A sequence of transformations is then performed iteratively on the simplex, aiming to decrease the function values at its vertices. Possible transformations include reflection, expansion, contraction, and shrinking. We refer readers to the original paper [13] for more details about these transformations. The above process terminates when the sample standard deviation of the function values of the current simplex fall below a predefined threshold.

In our context, the initial simplex has five vertices. Each vertex consists of four values assigned to the four LSTM hyperparameters. One of the five vertices is so called the predefined vertex, and it consists of four default values separately assigned to the four LSTM hyperparameters. The remaining four vertices are automatically determined by NMM in a deterministic way. In other words, NMM always produces the same four vertices given a predefined vertex. Note that the term "vertex" and "hyperparameter setting" are interchangeable. In this paper, the function evaluation is to derive the prediction error introduced by an LSTM model trained with a certain dataset under a specific hyperparameter setting. If the prediction error of an LSTM model is no larger than a predefined threshold, NMM terminates its searching processing.

## 3      Related Work

Traffic prediction approaches introduced in the past two decades can be classified into two categories: parametric approaches and nonparametric approaches. In parametric approaches, a model structure needs to be determined beforehand based on some theoretical assumptions. The ARIMA model is a typical and widely used parametric approach [14]. ARIMA is designed to fit time series data so as to predict future data points in the time series. Many ARIMA-based approaches were also introduced to improve prediction accuracy, including [15-17].

Different from parametric approaches, nonparametric approaches do not require a predefined model structure. There is no need to make assumptions about the mapping function. Typical examples include k-NN, ANN, RNN, hybrid approaches, etc. Le et al. [18] addressed traffic speed prediction using big traffic data obtained from static sensors and proposed local Gaussian Processes to learn and make predictions for correlated subsets of data. Jiang and Fei [19] introduced a data-driven vehicle speed prediction method based on Hidden Markov models. However, these two approaches focus on predicting traffic on a road section or a small region. They might be difficult to use in large-scale transportation networks.

Ma et al. [20] used deep learning theory to predict traffic congestion evolution in large-scale transportation networks. Furthermore, Ma et al. [21] predicted traffic speed in large-scale transportation networks by representing traffic as images and employing convolutional neural networks to make prediction. However, both of these methods require the scale of the target transportation network to be fixed and specified in advance. Lee et al. introduced DALC [22] to predict traffic speed of each individual detector in large-scale transportation networks based on LSTM. However, DALC only focuses on auto-tuning two LSTM hyperparameters, i.e., the number of hidden layers and epochs for each detector of the target transportation network.

Different from these methods, DistPre proposed in this paper is designed in an incremental manner. DistPre can handle an increasing number of detectors on the fly without pre-fixing the scale of the target transportation network, and it is able to automatically tune more LSTM hyperparameters for each detector if needed. These practical features make DistPre an ideal solution for providing fine-grained traffic speed prediction for large-scale and growing transportation networks.

## 4      The details of DistPre

The architecture of DistPre consists of a master node and a set of worker nodes. The master node decides when it is necessary to customize an LSTM model for each detector in the target transportation networks. Each worker node waits for an instruction from the master node and conducts the required LSTM customization process for a given detector upon request.

Algorithm 1 illustrates the algorithm of DistPre running on the master node. Let $G = \{D_1, D_2, ..., D_x\}$ be a list of detectors that already have their own LSTMs customized by DistPre. It is clear that $G$ is empty before DistPre is employed and launched.

Whenever DistPre encounters an unprocessed detector (denoted by $U_i$) in the target transportation networks, the master node first normalizes $L_i$, which is a list of traffic-speed values previously observed by $U_i$. Note that $L_i = \{v_{i,1}, v_{i,2}, \dots, v_{i,T}\}$ where $v_{i,t}$ is the traffic-speed value observed by $U_i$ at time point $t$, $t = 1, 2, \dots, T$. The normalization is to divide $v_{i,t}$ by $f$ where $f$ is a predefined fixed value (e.g., 70 to represent the speed limit in mph). The normalized $L_i$, denoted by $N_i$, will be $\{n_{i,1}, n_{i,2}, \dots, n_{i,T}\}$ where $n_{i,t} = \frac{v_{i,t}}{f}$.

The master node decides whether to customize an LSTM model for $U_i$ or not by sequentially comparing $U_i$ with every detector (denoted by $D_j$, $j = 1, 2, \dots, x$) in $G$ in terms of their normalized traffic-speed pattern based on the following equation:

$$\text{AARD}_{i,j} = \frac{1}{T} \cdot \sum_{t=1}^{T} \frac{|n_{i,t} - n_{j,t}|}{n_{i,t}} \tag{1}$$

where $\text{AARD}_{i,j}$ is the average absolute relative difference between the traffic-speed patterns collected by $U_i$ and $D_j$, and $n_{j,t}$ is the normalized traffic-speed value collected by $D_j$ at time point $t$, implying that $n_{j,t} = \frac{v_{j,t}}{f}$. If $\text{AARD}_{i,j}$ is less than a predefined threshold $thd_{AARD}$ (implying that $U_i$ and $D_j$ observe a similar traffic-speed pattern), the master node directly shares the LSTM of $D_j$ with $U_i$ (see lines 7 to 10 of the algorithm).

---

**Input:** An unprocessed detector $U_i$
**Output:** A decision to customize an LSTM model for $U_i$ or to share an LSTM model with $U_i$
**Procedure:**

| | |
|---|---|
| 1 | Let $M$ be a boolean variable and let $M$ be false; |
| 2 | Let $G = \{D_1, D_2, \dots, D_x\}$ be a list of detectors having their own LSTMs customized by DistPre; |
| 3 | Let $L_i = \{v_{i,1}, v_{i,2}, \dots, v_{i,T}\}$ be a list of traffic-speed values previously observed by $U_i$; |
| 4 | Normalize $L_i$ into $N_i$ by dividing each value in $L_i$ by $f$; |
| 5 | **for** $j = 1$ to $x\{$  // $x$ is the total number of detectors in $G$; |
| 6 |     Calculate $\text{AARD}_{i,j}$ based on Equation 1; |
| 7 |     **if** $\text{AARD}_{i,j} < thd_{AARD}\{$ |
| 8 |         Share the LSTM model of $D_j$ with $U_i$; |
| 9 |         Let $M$ be true; |
| 10 |         break;$\}\}$ |
| 11 | **if** $M$==false $\{$ |
| 12 |     Request an available worker node to customize an LSTM model for $U_i$; |
| 13 |     Append $U_i$ to the end of $G;\}$ |

**Algorithm 1.** The LSTM auto-tuning and sharing algorithm performed by the master node.

However, if the master node is unable to find any detector that has observed a similar traffic-speed pattern with $U_i$ (i.e., line 11 holds), the master node requests an available worker node from the cluster to customize an LSTM model for $U_i$, and then appends $U_i$ to the end of $G$ to indicate that $U_i$ will have its own LSTM model customized by DistPre. Based on how each detector is appended to $G$, it is clear that every detector in $G$ must have observed a distinct traffic-speed pattern.

On the other hand, whenever a worker node receives an LSTM customization request for $U_i$ from the master node, it utilizes NMM to automatically find appropriate values for the four abovementioned hyperparameters by using the following initial hyperparameter setting as the predefined vertex:

$$<R_{Learn} = 0.01, N_{Layer} = 1, N_{Unit} = 2, ep = 100>$$

Note that the predefined vertex consists of four low hyperparameter values. The goal is to enable NMM to start with a simple LSTM model since such a model introduces less computational cost than a more complex LSTM model.

When the worker node finds a hyperparameter setting which enables the corresponding LSTM to reach the required prediction accuracy for $U_i$ (i.e., the corresponding AARE value calculated based on Equation 2 is lower than or equal to a predefined threshold $thd_{AARE}$), the worker node terminates the customization process and outputs the LSTM model to be the LSTM model of $U_i$.

$$\text{AARE} = \frac{1}{W} \cdot \sum_{\omega=1}^{W} \frac{|s_\omega - \widehat{s_\omega}|}{s_\omega} \qquad (2)$$

Note that, in Equation 2, $W$ is the total number of data points considered for comparison, $\omega$ is the index of a data point, $s_\omega$ is the actual traffic-speed value at $\omega$, and $\widehat{s_\omega}$ is the forecast traffic-speed value at $\omega$.

## 5    Performance Evaluation

To evaluate DistPre, we chose freeway I5-N as our target transportation network. I5-N is a major route from the Mexico-United States border to Oregon with a total length of 796.432 miles. In our experiment, DistPre incrementally provides its LSTM customization and sharing service until the 110 detectors that are deployed on I5-N are completely covered. Note that the distance between two consecutive detectors is around 5 km. We crawled the traffic data collected by each of these 110 detectors for six continuous working days from the Caltrans performance measurement system [23], which is a database of traffic data collected by detectors placed on state highways throughout California. The traffic data of each detector is then split into a training dataset (the first 5 days) and a testing dataset (the last day). Due to the fact that all the traffic data is aggregated at 5-minute intervals, DistPre follows the same interval for prediction.

In this experiment, DistPre was deployed on a cluster running Apache Hadoop YARN 2.2.0 [24]. The cluster consists of one master node and 30 worker nodes. Each node runs Ubuntu 12.04.1 LTS with 2 CPU cores, 2GB of RAM, and 100GB of storage. As mentioned earlier, four LSTM hyperparameters are considered to be auto-tuned by DistPre. Table 1 lists the domain of these LSTM hyperparameters. For each hyperparameter, we choose a range of values for NMM to conduct its search process. Note that the maximum value for each hyperparameter was determined according to our previous experience.

**Table 1.** Four LSTM hyperparameters and their domains used by DistPre.

| Hyperparameter | Domain | Description |
|---|---|---|
| $R_{Learn}$ | [0.01, 0.2] | Discrete with step=0.01 |
| $N_{Layer}$ | [1, 10] | Discrete with step=1 |
| $N_{IInit}$ | [2, 40] | Discrete with step=2 |
| $ep$ | [100, 1000] | Discrete with step=20 |

The goal of this experiment is to study the impact of the LSTM sharing function and the number of worker nodes on the performance of DistPre. To this aim, the four cases listed in Table 2 were designed. In Case 1, we allowed only one worker node of the cluster to support the operation of DistPre. In addition, we disabled the LSTM sharing function of DistPre. In other words, each detector always gets its own LSTM model, and all the LSTM customizations are sequentially performed by the only worker node. In Case 2, we still limited only one worker node to support DistPre, but we enabled the LSTM sharing function. Therefore, detectors were able to share an LSTM model if they observed similar traffic patterns. In Cases 3 and 4, we increased the number of worker nodes to 30. However, we disabled and enabled the LSTM sharing function in Cases 3 and 4, respectively.

**Table 2.** The details of the four cases.

| Case No. | Number of worker nodes involved | The LSTM sharing function |
|---|---|---|
| 1 | 1 | Disabled |
| 2 | 1 | Enabled |
| 3 | 30 | Disabled |
| 4 | 30 | Enabled |

Note that $thd_{AARD} = 0.1$ and $thd_{AARE} = 0.05$ in all the cases. If two detectors have 90% similarity in their monitored traffic-speed patterns, we consider that they have similar patterns. This is why we set $thd_{AARD}$ to be 0.1. The same reason for $thd_{AARE}$: We consider that it is satisfactory if a detector is able to provide 95% prediction accuracy. This is why we set $thd_{AARE}$ to be 0.05. Note that these two thresholds are configurable if one wants to achieve a different level of prediction accuracy. The following five performance metrics are chosen in this experiment:

1. Total LSTM customization duration (TLCD). This is the time period starting when DistPre is launched and ending when all the 110 detectors obtain their LSTM models. Apparently, if TLCD is short, it means that DistPre is time efficient.

2. The total number of LSTMs generated by DistPre over time.

3. Average AARE, calculated as below:

$$\text{Average AARE} = \frac{\sum_{r=1}^{Z} \text{AARE}_r}{Z} \qquad (3)$$

where $\text{AARE}_r$ is the AARE value associated with the LSTM model of detector $r$, where $r = 1, 2, \ldots, Z$, and $Z$ is the total number of the detectors in the tar-get transportation networks. Note that $AARE_r$ is calculated based on Equation 2 and $Z$ equals to 110 in this experiment.

4. Average AAE, calculated as below:

$$\text{Average AAE} = \frac{\sum_{r=1}^{Z} \text{AAE}_r}{Z} \tag{4}$$

where $AAE_r$ is the average absolute error (AAE) value associated with the LSTM model of detector $r$, and $AAE_r$ is defined as $\frac{1}{W} \cdot \sum_{\omega=1}^{W} |s_{r,\omega} - \widehat{s_{r,\omega}}|$ [25]. A low AAE value implies that the forecast values are close to the actual values.

5. Average RMSE, calculated as below:

$$\text{Average RMSE} = \frac{\sum_{r=1}^{Z} \text{RMSE}_r}{Z} \tag{5}$$

where $RMSE_r$ is the root mean square error associated with the LSTM model of detector $r$, and $RMSE_r$ is defined as $\sqrt{\frac{1}{W} \cdot \sum_{\omega=1}^{W} (s_{r,\omega} - \widehat{s_{r,\omega}})^2}$ [25]. A low RMSE value suggests that the forecast values are close to the actual values.

Fig. 2 shows the TLCD results of DistPre in the four cases. Case 1 leads to the longest TLCD, which is around 3144 minutes. This is because only one worker node was employed to customize an LSTM model for each individual detector in Case 1. We can see that TLCD is significantly reduced in Case 2. The required TLCD is reduced by 81.46% (= $\frac{3144-583}{3144}$) from Case 1 to Case 2, implying that enabling detectors to share their LSTM models greatly reduces the number of times LSTM models need to be customized, even though there is only one worker node supporting the operation of DistPre.

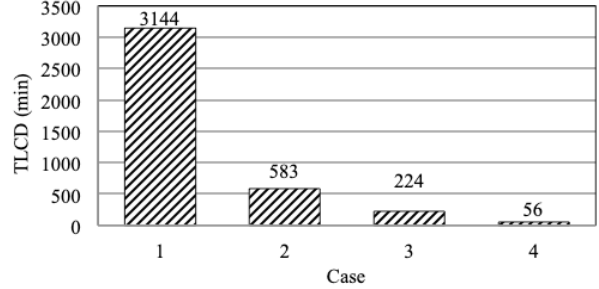

**Fig. 2.** The TLCD performance of DistPre in the four cases. Note that TLCD represents the time period from when DistPre is launched to when all the 110 detectors of I5-N obtain LSTM models from DistPre.

When 30 worker nodes are used by DistPre and the sharing function is disabled, i.e., Case 3, the required TLCD is reduced to 224 min, meaning that the distributed and parallel processing further improves the performance of DistPre, even when com-pared to Case 2 (single worker node, LSTM model sharing enabled). By further enabling the sharing function, i.e., Case 4, the total time duration drops to only 56 minutes. The reduction is around 75% (= $\frac{224-56}{224}$) compared with Case 3, and 98% (= $\frac{3144-56}{3144}$) compared to Case 1. This great performance improvement is mainly due to two factors. Firstly, by means of DistPre, only 31 out of the 110 detectors require a customized LSTM. Secondly, the work of LSTM customization is distributed to 30 worker nodes.

Altogether, the above results demonstrate that DistPre is able to provide the LSTM customization service in a time-efficient and scalable way for detectors in large-scale transportation networks. This feature is very important since large-scale transportation networks usually contain numerous amounts of detectors and the amounts may keep increasing. Furthermore, note that the number of worker nodes could be increased even further to handle even larger transportation networks when needed.

Fig. 3 illustrates the number of LSTM models customized by DistPre over time, i.e., as new detectors are processed. We can see that Case 1 and Case 3 have identical results: Whenever DistPre processed a new unknown detector, one more LSTM model is customized. The reason is that the LSTM sharing function is disabled in both cases, so every detector always gets its own customized LSTM model from DistPre. On the other hand, in Case 2 and Case 4, there is no one-to-one relationship between the number of LSTM models customized and the number of detectors processed by DistPre. When DistPre processed a new unknown detector, the number of customized LSTM models did not always increase due to the LSTM sharing function. In fact, when all the 110 detectors were processed by DistPre, only 31 LSTM models were generated and customized by DistPre. This also explains why DistPre in Case 2 and Case 4 have shorter TLCD than DistPre in Case 1 and Case 3, respectively.

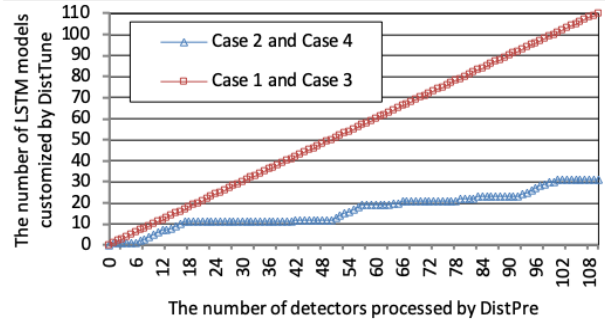

**Fig. 3.** The number of LSTM models customized by DistPre versus with the number of detectors processed by DistPre.

From the perspective of prediction performance, both Case 1 and Case 3 have the same results when it comes to average AARE, average AAE, and average RMSE as shown in Figs. 4, 5, and 6, respectively. The main reason is that the algorithm of NMM is deterministic. No matter which worker node executes NMM for the same detector, the result is always the same. Due to the same reason, the prediction accura-cy results in Case 2 and Case 4 are identical, but they are both lower than those in Case 1 and Case 3. This is because not all the detectors in Case 2 and Case 4 have customized LSTMs that perfectly fit their training data. Nevertheless, the average AARE values in Case 2 and Case 4 still satisfy our requirement since they are both lower than the predefined $thd_{AARE}$ (i.e., 0.05).

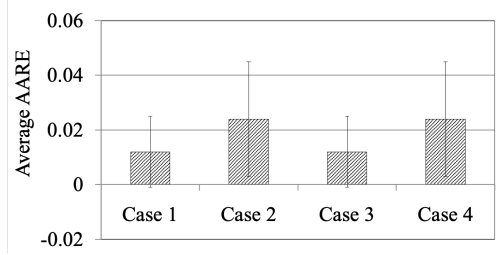

**Fig. 4.** The average AARE results in four cases.

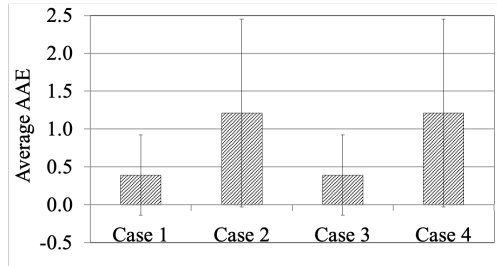

**Fig. 5.** The average AAE results in four cases.

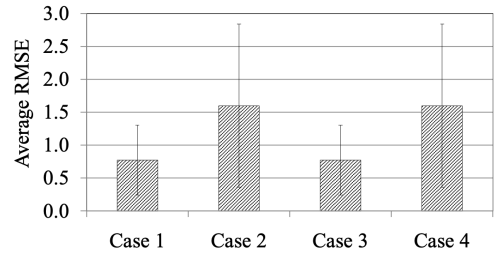

**Fig. 6.** The average RMSE results in four cases.

By combining all the above results, we can see that DistPre in Case 4 provides the best trade-off between time efficiency and prediction accuracy. Employing LSTM sharing and parallel processing for LSTM customization enables DistPre to provide fine-grained, accurate, and efficient traffic speed prediction for large-scale transportation networks.

# 6      Conclusion and Future Work

In this paper, we have introduced DistPre, a distributed scheme to achieve fine-grained, accurate, and efficient traffic speed prediction for a large amount of detectors deployed in large-scale transportation networks. DistPre automatically customizes an LSTM models with an appropriate hyperparameter setting for a detector based on NMM. By enabling any trained LSTM model to be shared between different detectors that all observe similar traffic-speed patterns, DistPre enables fine-grained and time-efficient traffic speed prediction in large-scale transportation networks. The required LSTM customization time does not proportionally increase when the number of detectors handled by DistPre increases. Our experiments based on real traffic data, collected by the Caltrans performance measurement system, demonstrate great performance of DistPre in both prediction accuracy and time efficiency.

As future work, we plan to extend DistPre and improve its performance by taking continuous monitoring and LSTM re-customization into account such that any detector is able to keep providing high prediction accuracy under any circumstances. In addition, we would like to investigate how DistPre can take advantage of the eX$^3$ HPC cluster [26] to further improve the performance of DistPre by investigating appropriate scheduling approaches such as [27-28].

**Acknowledgments.** This work was supported by the project eX$^3$ - *Experimental Infrastructure for Exploration of Exascale Computing* funded by the Research Council of Norway under contract 270053 and the scholarship under project number 80430060 supported by Norwegian University of Science and Technology.

# References

1. Ahmed, M. S. Cook, A. R.: Analysis of freeway traffic time-series data by using Box-Jenkins techniques, 722 (1979).
2. Davis, G.A., Nihan, N.L.: Nonparametric regression and short-term freeway traffic forecasting. Journal of Transportation Engineering 117(2), 178–188 (1991). DOI:10.1061/(ASCE)0733-947X(1991)117:2(178)
3. Bustillos B., Chiu, Y. C.: Real-time freeway-experienced travel time prediction using N-curve and k nearest neighbor methods. Transportation Research Record: Journal of the Transportation Research Board 2243, 127–137 (2011). DOI:10.3141/2243-15
4. Chan, K.Y., Dillon, T.S., Singh, J., Chang, E.: Neural-network-based models for short-term traffic flow forecasting using a hybrid exponential smoothing and Levenberg–Marquardt algorithm. IEEE Transaction Intelligence Transportation Systems 13(2), 644–654 (2012). DOI: 10.1109/TITS.2011.2174051
5. van Lint, J.W.C., Hoogendoorn, S.P., van Zuylen, H.J.: Freeway Travel Time Prediction with State-Space Neural Networks: Modeling State-Space Dynamics with Recurrent Neural Networks. Transportation Research Record 1811(1), 30–39 (2002). DOI: 10.3141/1811-04
6. Hochreiter, S., Schmidhuber J.: Long short-term memory. Neural computation 9(8), 1735–1780 (1997). DOI: 10.1162/neco.1997.9.8.1735

7. Ma, X., Tao, Z., Wang, Y., Yu, H., Wang, Y.: Long short-term memory neural network for traffic speed prediction using remote microwave sensor data. Transportation Research Part C: Emerging Technologies 54, 187–197 (2015). DOI: 10.1016/j.trc.2015.03.014

8. Yu, R., Li, Y., Shahabi, C., Demiryurek, U., Liu, Y.: Deep learning: A generic approach for extreme condition traffic forecasting. In Proceedings of the 2017 SIAM International Conference on Data Mining. Society for Industrial and Applied Mathematics, pp. 777–785. (2017). DOI: 10.1137/1.9781611974973

9. Zhao, Z., Chen, W., Wu, X., Chen, P.C., Liu, J.: LSTM network: a deep learning approach for short-term traffic forecast. IET Intelligent Transport Systems 11(2), 68–75 (2017). DOI: 10.1049/iet-its.2016.0208

10. Interstate 5 in California, https://en.wikipedia.org/wiki/Interstate_5_in_California, last accessed 2020/05/31.

11. Nelder, J.A., Mead, R.: A simplex method for function minimization. The computer journal 7(4), 308–313 (1965). DOI: 10.1093/comjnl/7.4.308

12. Hochreiter, S.: The vanishing gradient problem during learning recurrent neural nets and problem solutions. International Journal of Uncertainty, Fuzziness and Knowledge-Based Systems 6(2), 107–116 (1998). DOI: 10.1142/S0218488598000094

13. Singer, S., Nelder, J.: Nelder-mead algorithm. Scholarpedia 4(7), 2928 (2009).

14. Box, G.E., Jenkins, G.M., Reinsel G.C.: Time series analysis: forecasting and control. John Wiley & Sons (2015).

15. Lee, S., Fambro, D.: Application of subset autoregressive integrated moving average model for short-term freeway traffic volume forecasting. Transportation Research Record: Journal of the Transportation Research Board 1678, 179–188 (1999). DOI: 10.3141/1678-22

16. Williams, B.: Multivariate vehicular traffic flow prediction: evaluation of ARIMAX modeling. Transportation Research Record: Journal of the Transportation Research Board (1776), 194–200 (2001). DOI: 10.3141/1776-25

17. Williams, B.M., Hoel, L.A.: Modeling and forecasting vehicular traffic flow as a seasonal ARIMA process: Theoretical basis and empirical results. Journal of transportation engineering 129(6), 664–672 (2003). DOI: 10.1061/(ASCE)0733-947X(2003)129:6(664)

18. Le, T.V., Oentaryo, R., Liu, S., Lau, H.C.: Local Gaussian processes for efficient fine-grained traffic speed prediction. IEEE Transactions on Big Data 3(2), 194–207 (2017). DOI: 10.1109/TBDATA.2016.2620488

19. Jiang B., Fei, Y.: Vehicle speed prediction by two-level data driven models in vehicular networks. IEEE Transactions on Intelligent Transportation Systems 18(7), 1793–1801 (2017). DOI: 10.1109/TITS.2016.2620498

20. Ma, X., Yu, H., Wang, Y., Wang, Y.: Large-scale transportation network congestion evolution prediction using deep learning theory. PloS one 10(3), e0119044 (2015). DOI: 10.1371/journal.pone.0119044

21. Ma, X., Dai, Z., He, Z., Ma, J., Wang, Y., Wang, Y.: Learning traffic as images: a deep convolutional neural network for large-scale transportation network speed prediction. Sensors 17(4), 818, (2017). DOI: 10.3390/s17040818

22. Lee, M.-C., Lin, J.-C.: DALC: Distributed Automatic LSTM Customization for Fine-Grained Traffic Speed Prediction. In Proceedings of the 34th International Conference on Advanced Information Networking and Applications, pp. 164–175. (2020) Springer, Cham, https://arxiv.org/abs/2001.09821. DOI: 10.1007/978-3-030-44041-1_15

23. PeMS, http://pems.dot.ca.gov/, last accessed 2020/05/31.

24. Apache Hadoop YARN, https://hadoop.apache.org/docs/current/hadoop-yarn/hadoop-yarn-site/YARN.html, last accessed 2020/05/31.

25. Zou, N., Wang, J., Chang, G.L., Paracha, J.: Application of advanced traffic information systems: field test of a travel-time prediction system with widely spaced detectors. Transportation Research Record 2129(1), 62–72 (2009). DOI: 10.3141/2129-08

26. Simula Research Laboratory eX$^3$ research cluster, https://www.ex3.simula.no, last accessed 2020/05/31.

27. Lee, M.-C., Lin, J.-C., Yahyapour, R.: Hybrid Job-driven Scheduling for Virtual MapReduce Clusters. IEEE Transactions on Parallel and Distributed Systems (TPDS) 27(6), 1687–1699 (2016). DOI: 10.1109/TPDS.2015.2463817

28. Lin, J.-C., Lee, M.-C.: Performance Evaluation of Job Schedulers under Hadoop YARN. Concurrency and Computation: Practice and Experience (CCPE) 28(9), 2711–2728 (2016). DOI: 10.1002/cpe.3736